\documentclass{article} 
\makeatletter
\providecommand{\@LN}[2]{}
\makeatother
\usepackage{iclr2026_conference,times}

\usepackage{amsmath,amsfonts,bm}

\def\eqref#1{equation~\ref{#1}}

\def\1{\bm{1}}

\DeclareMathAlphabet{\mathsfit}{\encodingdefault}{\sfdefault}{m}{sl}
\SetMathAlphabet{\mathsfit}{bold}{\encodingdefault}{\sfdefault}{bx}{n}

\usepackage{hyperref}
\usepackage{url}
\usepackage{multirow}
\usepackage{arydshln}
\usepackage{graphicx}  
\usepackage{booktabs} 
\usepackage{diagbox}
\usepackage{graphicx}
\usepackage{colortbl}
\usepackage{tcolorbox}
\usepackage{booktabs}
\usepackage{multirow}
\usepackage{xspace}
\usepackage{array}
\usepackage{subcaption}
\usepackage{wrapfig}
\usepackage{enumitem}
\usepackage{xcolor}      
\usepackage{colortbl}    
\usepackage{pifont}      
\usepackage{fontawesome5}
\usepackage{hhline}

\title{Training LLM Agents to Automate LLM Post-Training with RSI-Dojo}

\author{%
\begin{tabular}[t]{@{}l@{}}
\textbf{Shuzheng Gao\textsuperscript{1} \quad
Zhaojian Yu\textsuperscript{2} \quad
Penghao Yin\textsuperscript{3} \quad
Yuling Shi\textsuperscript{4} \quad
Kai Cai\textsuperscript{2}} \\
\textbf{Shilin He\textsuperscript{2} \quad
Michael R. Lyu\textsuperscript{1}} \\[0.2cm]
{\normalfont\textsuperscript{1}The Chinese University of Hong Kong} \quad
{\normalfont\textsuperscript{2}ByteDance \quad
\textsuperscript{3}Tsinghua University} \\
{\normalfont\textsuperscript{4}Shanghai Jiao Tong University}
\end{tabular}%
}

\newcommand{\tool}{RSI-Dojo\xspace}
\newcommand{\builder}{RSI-Craft\xspace}

\iclrfinalcopy 
\begin{document}

\maketitle

\footnotetext[1]{\small Correspondence: Shuzheng Gao {\href{mailto:szgao23@cse.cuhk.edu.hk}{szgao23@cse.cuhk.edu.hk}}, 
Shilin He {\href{mailto:heshilin@bytedance.com}{heshilin@bytedance.com}}}

\begin{abstract}
Recursive Self-Improvement (RSI) aims to enable AI systems to autonomously build and improve AI systems themselves, including models, harnesses, and underlying infrastructure. Within this broader agenda, a central capability is enabling agents to train and improve base models. However, existing studies on automated model training primarily focused on workflow- or harness-level optimizations, rather than internalizing the ability to conduct post-training into the model itself. Developing this capability requires executable model training environments that can produce high-quality trajectories demonstrating how to select data, train models, diagnose failures, and iteratively improve results. In this paper, we introduce \tool{}, a collection of executable model training environments designed to teach LLM agents to conduct post-training. To construct \tool{}, we develop \builder{}, an automatic pipeline that converts human research artifacts into executable model training environments. \builder{} automatically collects research papers from top venues, instantiates datasets and metrics into a unified interface, and reproduces evaluation dependencies. It then filters tasks by data size, execution reliability, and discriminability, yielding \tool{} with 281 seed model training tasks. Building on \tool{}, we employ a trajectory collection method with an improved harness to collect high-quality model-training trajectories from frontier models, then retain examples through rubric-based and outcome-based filtering. We fine-tune Qwen3.5-9B and Qwen3.6-35B-A3B on these trajectories. The resulting RSI-Dojo-9B and RSI-Dojo-35B-A3B improve over their base models by 8.74 and 16.87 PostTrainBench points, respectively. RSI-Dojo-35B-A3B reaches 29.95 on PostTrainBench and 53.29 on PostTrainBench-Lite, matching the performance of frontier models such as GLM-5.2 and GPT 5.5.
\end{abstract}

\begin{quote}
\small
\begin{tabular}{@{}l@{\hspace{0.5em}}l@{}}
\faGithub & \textbf{Github Repo}: \url{https://github.com/shuzhenggao/rsi-dojo}\\
\raisebox{-0.15em}{\includegraphics[height=0.95em]{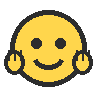}} & \textbf{Data}: \url{https://huggingface.co/datasets/szgao/RSI-Dojo}\\
\faBookOpen & \textbf{Homepage}: \url{https://shuzhenggao.github.io/rsidojo/}
\end{tabular}
\end{quote}

\begin{figure}[h]
    \centering
    \includegraphics[width=\linewidth]{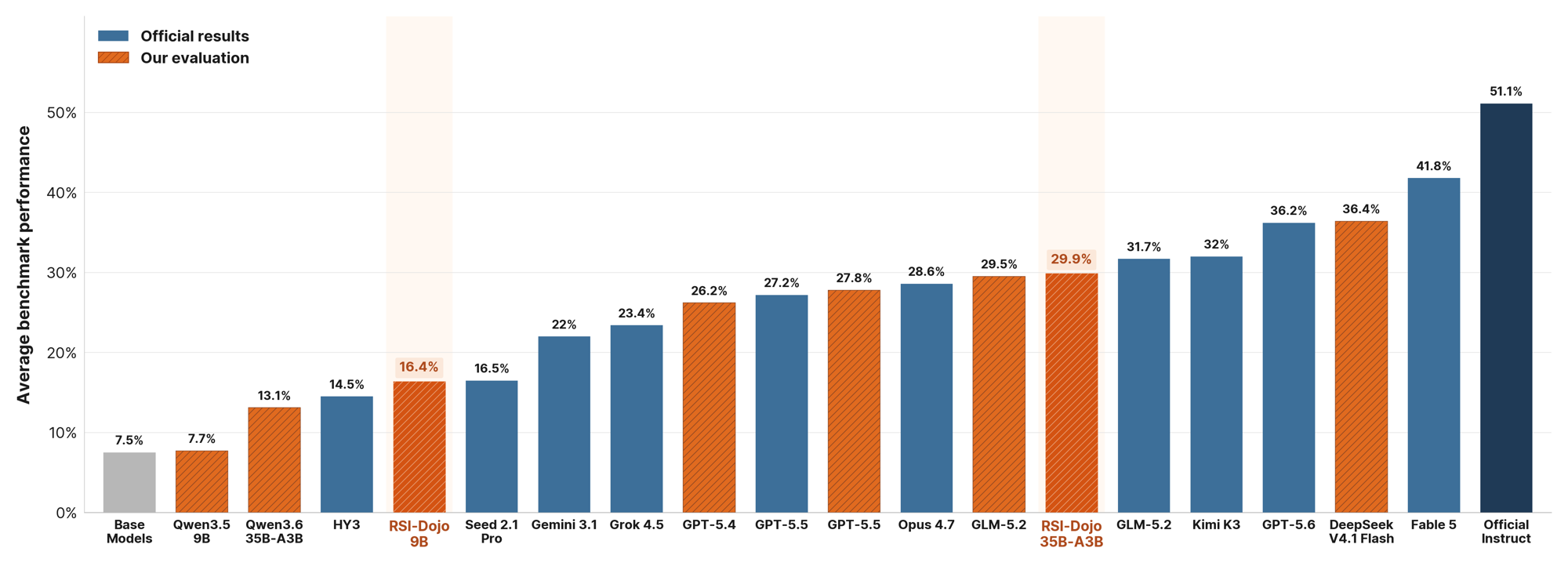}
    \caption{Performance of RSI-Dojo-35B-A3B and RSI-Dojo-9B on PostTrainBench. Official results are taken from PostTrainBench v1.1 or the corresponding technical report. Our evaluation results are reproduced under the same harness and environment settings (Details are provided in Section~\ref{sec:evaluation-environments}).}
    \label{fig:intro-scaling}
\end{figure}

\section{Introduction}

As LLM agents are increasingly deployed to tackle complex and long-horizon tasks~\cite{yang2026programbench,dong2026longhorizon,gao2026swemem}, the use of AI systems to build and improve AI systems has become central to the prospect of recursive self-improvement (RSI) \cite{yang2026selfimprovementlargelanguagemodels}. Recent benchmarks have begun to make this ambition measurable by evaluating agents’ ability to design and execute reinforcement-learning pipelines \cite{chen2026agent2rlbench}, automate iterative data curation against training and evaluation feedback \cite{kang2026curationbench}, and conduct the whole post-training process \cite{rank2026posttrainbenchllmagentsautomate}. To strengthen these capabilities, recent work has explored improving agents' discovery and curation of training data \cite{wang2026datamaster}, structuring their interactions with tools through explicit workflows and rules \cite{yu2026autotrainess}, and adapting model-specific harnesses from execution feedback \cite{zhang2026selfharness}.

Although these methods have achieved promising results, they mainly use harness- and workflow-level optimization to teach agents how to design data and training recipes, analyze results, and literately optimize the performance. However, the capability to build and improve AI systems is not internalized into the model itself. A central challenge is how to construct suitable tasks and environments, and how to collect high-quality training trajectories to teach the model how to train models. Unlike SWE tasks which can leverage many real developer issues with repository context and test suites \cite{jimenez2024swebench,pan2025swegym}, or MLE tasks which can reuse Kaggle competitions with established datasets, evaluation metrics, and competitive baselines \cite{chan2025mlebench}, how to automatically construct training tasks and environments for LLM post-training at scale remains unexplored.

To address this gap, we introduce \builder{}, an automatic pipeline for constructing executable model training environments from human research artifacts. \builder{} starts from evaluation platforms and top-venue papers and uses paper selection, dataset instantiation, and evaluation reproduction to recover tasks that can be executed in a unified post-training workflow. It then normalizes dataset variants and filters tasks by data size, execution reliability, and discriminability. Based on the above process, it constructs \tool{}, a suite of executable environments for automated post-training with 281 seed tasks.

Building on \tool{}, we collect model-training trajectories from frontier models and use them as supervision for training automated post-training agents. Since even frontier models still produce trajectories with recurring bad patterns during model training, we employ a trajectory collection method with an improved harness that adds targeted rollout-time guidance while preserving the end-to-end training loop. The collected trajectories then pass through a rubric-based quality filter to remove data with bad patterns and an outcome filter that checks whether the final submitted model yields a measurable task improvement. We use the retained trajectories to fine-tune Qwen3.5-9B and Qwen3.6-35B-A3B and obtain RSI-Dojo-9B and RSI-Dojo-35B-A3B. The two models score 16.46 and 29.95 on PostTrainBench, improving over their base models by 8.74 and 16.87 points, respectively. RSI-Dojo-35B-A3B reaches 29.95 on PostTrainBench and 53.29 on PostTrainBench-Lite, matching the performance of frontier models such as GLM-5.2 and GPT 5.5 under the same evaluation setting.

The main contributions of this paper are as follows:
\begin{itemize}[leftmargin=1.4em]
    \item We introduce \builder{}, an automatic pipeline that converts recent dataset-oriented research papers into executable post-training environments and constructs \tool{}.
    \item Based on \tool{}, we employ a trajectory collection method with an improved harness to collect high-quality trajectories from frontier models and retain those demonstrating strong process quality and measurable task improvement.
    \item We fine-tune Qwen3.5-9B and Qwen3.6-35B-A3B on the retained trajectories and obtain RSI-Dojo-9B and RSI-Dojo-35B-A3B, which achieves 16.46 and 29.95 on PostTrainBench, respectively.
\end{itemize}

\section{Related work}

\subsection{Recursive self-improvement and self-evolving agents}

Recursive self-improvement (RSI) studies AI systems that participate in improving systems \cite{ouyang2025kernelbench,chen2025cudallm,gao2025searchbased}, harnesses \cite{gao2025promptalchemist,zhang2026selfharness,wu2026harnessdev}, or base models \cite{yang2026selfimprovementlargelanguagemodels,yu2026autotrainess}.
Recent work has introduced a series of benchmarks for RSI-related problems. For example, Agent$^2$ RL-Bench \cite{chen2026agent2rlbench} evaluates whether agents can design, implement, debug, and execute reinforcement-learning post-training pipelines under a fixed workspace and budget. CurationBench \cite{kang2026curationbench} studies automated data curation, including discovering, filtering, and iterating on data resources. RSIBench-Data \cite{meng2026rsibenchdata} further focuses on data-centric research for RSI, where agents revise training-data strategies for a fixed target model and learn from checkpoint feedback. Evo-Bench \cite{huang2026evobench} evaluates intrinsic harness-evolving capability across Search, Office, and General domains while controlling for base model capabilities and task-specific overfitting. HarnessDev \cite{wu2026harnessdev} further decomposes harness development into Creation and Evolution, where it builds a runnable execution system from a minimal seed and keeps improving it using downstream feedback. 
Apart from those methods, another line of work improves the agent harness for RSI. Self-Harness \cite{zhang2026selfharness} treats the harness as the object of improvement through weakness mining, harness proposal, and regression-tested validation. Dream-RSI \cite{zheng2026dreamrsi} improves exploration by building replay simulators from discovery histories and using them to refine exploration policies before redeployment. Hyperagents \cite{zhang2026hyperagents} integrates task-solving and meta-level self-modification into a single editable program, allowing the improvement mechanism itself to evolve.  Other self-improvement methods use self-generated rationales or feedback to improve the performance \cite{zelikman2022star,madaan2023selfrefine,shinn2023reflexion,yuan2024selfrewarding}.

\subsection{Automated post-training}

PostTrainBench \cite{rank2026posttrainbenchllmagentsautomate} frames automated post-training as an agent task in which a base model, target benchmark, evaluation interface, and bounded compute budget are provided, and the agent must raise downstream performance through data discovery, recipe selection, training, and iteration. Recently, several works have improved model performance on PostTrainBench by adding structure to the agent's post-training loop. DataMaster \cite{wang2026datamaster} focuses on data-side search: it organizes alternative data-engineering attempts in a DataTree, shares discovered resources through a Data Pool, and records reusable evidence in Global Memory. AutoTrainess \cite{yu2026autotrainess} improves the agent-computer interface by exposing planning, data preparation, training, evaluation, and logging as explicit workflows and execution constraints. ANDES \cite{zhao2026andes} treats data synthesis as a plug-in agent skill, using a self-evolving World Tree and diagnostic reports to steer closed-loop data generation. RSI-Master \cite{du2026rsimaster} structures autonomous model-improvement experiments through an Experiment OS and reviewer-guided research orchestration, regularizing individual actions while encouraging exploration across research directions. G\"odel Forest \cite{zhao2026godelforest} organizes data-centric RSI as multiple co-evolving search trees, combining persistent local exploration with shared procedural memory to balance search depth and breadth. Harness design also interacts closely with post-training \cite{kim2026interplayharnessdesignposttraining}. 
These methods improve the workflow and agent harness for automated post-training tasks, while \tool{} constructs a scalable source of executable post-training tasks that enables collecting trajectories that can internalize the post-training capability into the base model.

\section{Methodology}

To train an agent for automated model training, we first introduce \builder{}, an automatic pipeline for constructing executable post-training environments from seed tasks selected from recent research papers and the InspectAI evaluation platform. \builder{} turns paper artifacts and existing task specifications into model training tasks and yields the \tool{}, which contains 281 tasks from 10 domains. Based on \tool{}, we employ a trajectory collection method with an improved harness to obtain high-quality agent trajectories and mitigate undesirable trajectory patterns. The collected trajectories then undergo quality filtering and are used to train the agent model, improving its post-training capabilities.

\begin{figure*}[t]
    \centering
    \includegraphics[width=\textwidth]{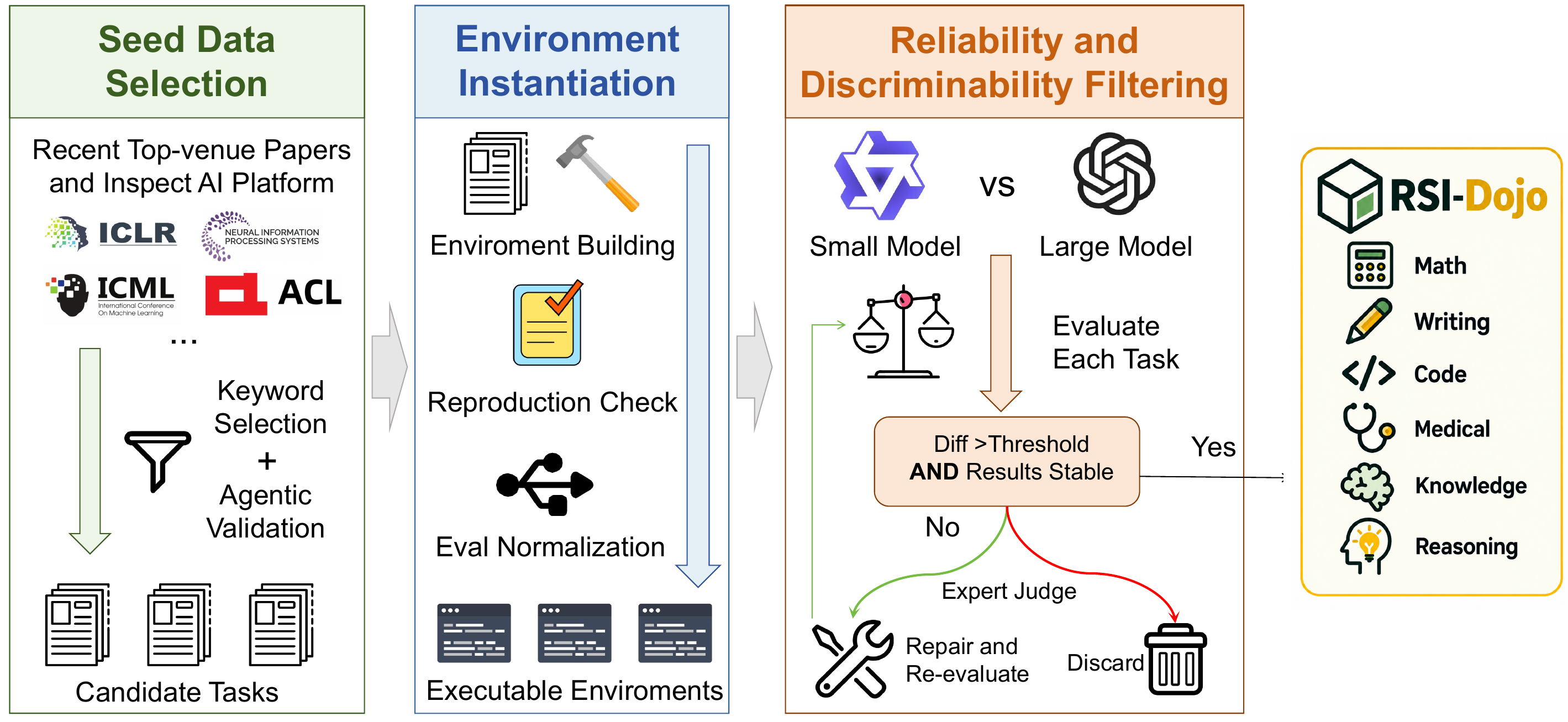}
    \caption{Overview of the \builder{} pipelinedojo. It automatically collects seed data from research papers and evaluation platforms, instantiates environments and metrics into a unified interface, and reproduces evaluation dependencies. It then filters tasks by execution reliability and discriminability and obtains \tool{}.}
    \label{fig:overview}
\end{figure*}

\subsection{RSI-Dojo  Construction}

\textbf{Seed data selection.} \builder{} starts by selecting seed tasks from two sources: recent research papers that may introduce datasets for evaluating the post-training performance of base models, and tasks available on the InspectAI evaluation platform. Specifically, it crawls papers from NeurIPS, ICML, ICLR, ACL, EMNLP, NAACL, and COLM over the last three years, and also imports candidate tasks from InspectAI. A keyword scanner then searches paper titles and abstracts for terms that indicate a dataset or benchmark. Each keyword-selected paper or imported InspectAI task is screened by an agent to determine whether it provides a real dataset or executable evaluation target, rather than only an evaluation method, analysis protocol, or task formulation without reusable data. Based on this procedure, \builder{} yields 547 candidate seed tasks for detailed inspection.

\textbf{Environment instantiation.} \builder{} then builds an evaluation environment for each selected task. Specifically, an agent locates the corresponding dataset and verifies that the data can be accessed. It also identifies the task definition, evaluation metric, and the code or protocol required to run the evaluation. For each dataset, \builder{} reproduces the evaluation environment by installing the required libraries and validating that the evaluation script or protocol can run successfully. Tasks that pass this execution check are then normalized into a unified evaluation interface, allowing the post-training agent to invoke different tasks through a consistent workflow. If a dataset contains multiple subsets with different metrics, these subsets are split into separate tasks. Tasks with fewer than 50 examples in total are discarded, while tasks with more than 300 examples are randomly downsampled to 300 examples to keep evaluation time tractable. This step produces 315 runnable initial task environments.

\textbf{Reliability and discriminability filtering.} The instantiated environments are then filtered for evaluation reliability and discriminability. The reliability filter removes tasks whose evaluation requires specialized infrastructure (i.e., kernel-relevant benchmarks) or yields unstable evaluation outcomes. This includes LLM-as-judge tasks whose repeated judgments vary substantially, as well as tasks whose metrics do not employ the model's generated output text. For the remaining tasks, the discriminability filter removes datasets that fail to separate frontier models from smaller models. Specifically, it evaluates the performance of Qwen3-4B-Base and GPT-5-mini on each task. A task is kept only when the score gap between the two models is at least 10 percentage points. Tasks where both models score above 80\% or both score below 20\% are also removed. Such tasks provide little usable signal for post-training.

\begin{wrapfigure}{r}{0.35\columnwidth}
    \vspace{-0.8\baselineskip}
    \centering
    \includegraphics[width=\linewidth]{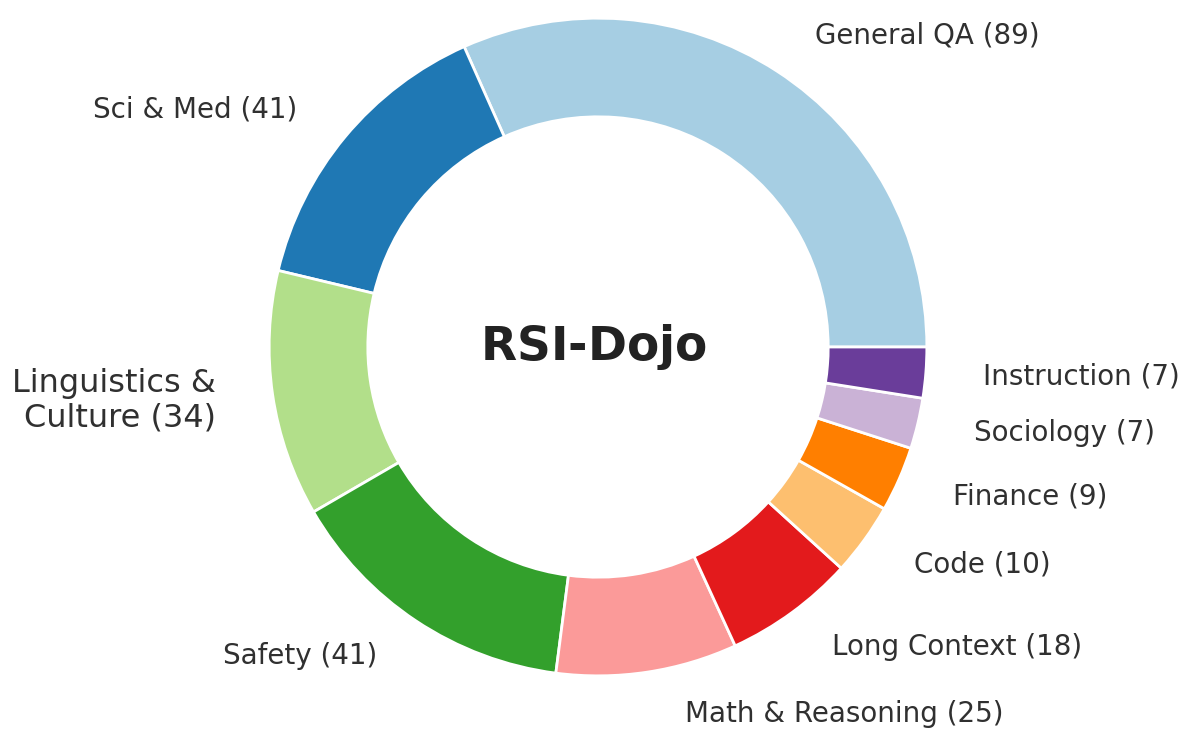}
    \caption{Domain distribution of the 281 \tool{} training tasks.}
    \label{fig:rsi-dojo-domain-distribution}
    \vspace{-0.6\baselineskip}
\end{wrapfigure}

\textbf{Dataset statistics.} The final \tool{} collection comprises 281  training environments spanning 10 domains, including General QA, Science \& Medicine, and Math \& Reasoning (as shwon in Figure~\ref{fig:rsi-dojo-domain-distribution}). General QA and Safety are the two largest categories, accounting for 31.7\% and 14.6\% of the collection, respectively, while Finance, Sociology, and Instruction Following form smaller specialized groups. It  combines broad domains that provide abundant post-training signal with less frequent domains that test whether agents can transfer their data selection, training-recipe construction, and evaluation strategies across tasks.

\subsection{Trajectory Collection with Improved Harness}

Based on \tool{}, we collect model-training trajectories from frontier models. However, direct rollouts from frontier models still exhibit limited effectiveness on automated post-training tasks. Even frontier models often produce undesirable patterns that degrade the quality of the collected training data. For example, an agent may perform multiple LoRA training runs on very small datasets while leaving the EOS configuration unchanged, causing an instruction-tuned model converted from a base model to fail to terminate generations properly. Some trajectories spend many rounds trying different recipes without analyzing concrete failure cases; as a result, the agent repeatedly explores alternative training settings without a clear direction for improvement. Other trajectories fail to submit the final model in the required format. To reduce these bad patterns, we employ a trajectory collection method with an improved harness. The base harness follows AutoTrainess \cite{yu2026autotrainess}, which provides task-specific guidance through \texttt{AGENTS.md}, \texttt{skills}, and \texttt{hooks} during the rollout process. We further augment this harness with additional guidance targeting the identified failure patterns. These instructions guide the agent to diagnose current model failures, choose suitable data and training recipes based on observed cases, check configuration details such as special tokens, and submit the final trained model in the expected format. 

\subsection{Trajectory Filtering and Model Training}

After data collection, we then filter low-quality trajectories for model training. We remove trajectories that contain the bad patterns described above, such as premature LoRA training with incorrect special-token handling, repeated recipe changes without case diagnosis, and failure to submit the final model in the required format. We also remove trajectories detected as cheating. Finally, we remove trajectories whose final model does not achieve a measurable improvement on the target task. 
The remaining trajectories are then used to train the agent model. 

\section{Experimental Setup}

\subsection{Benchmarks and Metrics}

We evaluate on PostTrainBench v1.1 \cite{rank2026posttrainbenchllmagentsautomate} and PostTrainBench-Lite. PostTrainBench evaluates automated post-training across four base models and seven target benchmarks. For each task, an agent receives a pretrained base model, an evaluation interface, one GPU, and a 10-hour budget, and must improve the target-benchmark performance through data construction, training, and iterative evaluation. PostTrainBench-Lite follows the evaluation setting introduced in the GPT-5.6 System Card \cite{openai2026gpt56systemcard}. It covers 12 combinations formed by three base models (Qwen3-4B-Base, Qwen3-1.7B-Base, and SmolLM3-3B-Base) and four target benchmarks (AIME 2025, BFCL, GSM8K, and HumanEval), while reducing the time budget to five hours. We use this setting for lower-cost comparisons and ablation studies.

We adopt the official evaluation metrics. For each agent, PostTrainBench evaluates a Cartesian product of base model $m$ and target benchmark $b$. Let $s_{m,b}$ denote the final benchmark score of the model produced by the agent on this point and $s^{\mathrm{base}}_{m,b}$ the corresponding pretrained base-model score. Following the official aggregation procedure, we use $\tilde{s}_{m,b}=\max(s_{m,b},s^{\mathrm{base}}_{m,b})$, so a submitted model that underperforms the base model receives the base score. The official score first applies benchmark-level weights and then averages across base models:
\[
\mathrm{Score}=\frac{1}{|\mathcal{M}|}\sum_{m\in\mathcal{M}}\sum_{b\in\mathcal{B}} w_b \tilde{s}_{m,b}.
\]
PostTrainBench-Lite instead normalizes each submitted model against its base and instruction-tuned reference models. For a Lite task with base model $m$ and target benchmark $b$, its reward is
\[
r_{m,b}=\operatorname{clamp}\left(
\frac{s^{\mathrm{trial}}_{m,b}-s^{\mathrm{base}}_{m,b}}
     {s^{\mathrm{instruct}}_{m,b}-s^{\mathrm{base}}_{m,b}},
0,1\right).
\]
Here, $s^{\mathrm{trial}}_{m,b}$ is the submitted model's score, $s^{\mathrm{base}}_{m,b}$ is the pretrained base model's score, and $s^{\mathrm{instruct}}_{m,b}$ is the corresponding instruction-tuned reference score. Matching the base model gives a score of 0, while matching or exceeding the instruction-tuned model gives a score of 1. Following the official protocol, trials judged to use invalid strategies, such as training on held-out evaluation data, receive a score of 0 \cite{openai2026gpt56systemcard}.



\subsection{Evaluation Models and Environments}
\label{sec:evaluation-environments}

We evaluate Qwen3.5-9B and Qwen3.6-35B-A3B before and after fine-tuning on \tool{} trajectories, and compare them with a broader set of frontier models. We evaluate GPT-5.5, GPT-5.4, GLM-5.2, and DeepSeek-V4.1-Flash ourselves under the same evaluation setting used for our trained models. We set the reasoning effort to \texttt{medium} for GPT-5.4 and GPT-5.5, and to \texttt{max} for GLM-5.2 and DeepSeek-V4.1-Flash. Results for the remaining reference models are taken from the official PostTrainBench leaderboard or their own technical reports. 

Our evaluation environment follows the PostTrainBench protocol with several infrastructure-specific adaptations. Due to hardware availability, we conducted the experiments on a less powerful GPU than the one required by the official benchmark. All agents use Codex CLI v0.142.4 with the standard AutoTrainess reprompting strategy and skills \cite{yu2026autotrainess}. During evaluation, we do not include the additional guidance introduced for trajectory collection. Codex exposes web search through the custom tool \texttt{web\_search}. This tool is unavailable to our self-hosted models, although they retain general network access. To provide sufficient training resources, we pre-download the datasets listed in the official PostTrainBench resource collection and the Hugging Face resources used by other models in traces published on the PostTrainBench website. This evaluation setup also helps prevent invalid strategies in PostTrainBench.

\subsection{Experiment Details}

During trajectory collection, we 
limit each rollout to five hours to control collection cost. After filtering, we retain 324 high-quality CLI sessions, which are split into 1,240 training trajectories. We disable the \texttt{web\_search} tool during collection to match the tool configuration used when evaluating our trained models. For each training task, we also pre-download suitable training resources to the local environment. After each rollout, we reconstruct training examples from the recorded Codex session by splitting at context compaction boundaries and excluding the reasoning summaries.  We apply rule-based conversions to custom tools such as \texttt{exec} and \texttt{wait}, mapping them to standard tool-call representations when possible and masking calls that cannot be converted. For \texttt{apply\_patch}, we use its open-source implementation to enable trained models to invoke this tool during inference. We train for four epochs with a learning rate that starts at $1\times10^{-5}$ and decreases to a minimum of $2\times10^{-6}$. 

\section{Experimental Results}

\subsection{Main Results}
\label{sec:main-results}

Table~\ref{tab:ptb-main-comparison} summarizes our main comparison setting on PostTrainBench and PostTrainBench-Lite. Based on the results, we achieve the following findings:

\begin{table*}[t]
\centering
\caption{Main comparison on PostTrainBench and PostTrainBench-Lite.}
\label{tab:ptb-main-comparison}
\setlength{\tabcolsep}{2.5pt}
\scriptsize
\resizebox{\textwidth}{!}{%
\begin{tabular}{lrrrrrrrrr}
\toprule
Model & \multicolumn{5}{c}{PostTrainBench} & \multicolumn{4}{c}{PostTrainBench-Lite} \\
\cmidrule(lr){2-6} \cmidrule(lr){7-10}
& Avg. & Qwen-1.7B & Qwen-4B & SmolLM3-3B & Gemma3-4B
& Avg. & Qwen-1.7B & Qwen-4B & SmolLM3-3B \\
\midrule
\multicolumn{10}{l}{\textit{Our Evaluation Results}} \\
Qwen3.5-9B & 7.72 & 6.65 & 15.09 & 4.52 & 4.60 & 0.56 & 0.00 & 1.67 & 0.00 \\
RSI-Dojo-9B & \textbf{16.46} & 15.05 & 21.48 & 14.65 & 14.66 & \textbf{15.71} & 11.47 & 15.67 & 19.99 \\
Qwen3.6-35B-A3B & 13.08 & 10.15 & 17.93 & 11.60 & 12.63 & 5.80 & 12.87 & 4.54 & 0.00 \\
RSI-Dojo-35B-A3B & \textbf{29.95} & 28.60 & 34.85 & 27.65 & 28.69 & \textbf{53.29} & 52.49 & 58.48 & 48.90 \\
\noalign{\vskip 2pt}
\cdashline{1-10}
\noalign{\vskip 2pt}
GPT-5.4 & 26.21 & 25.68 & 31.68 & 23.11 & 24.37 & 44.02 & 49.10 & 41.49 & 41.48 \\
GPT-5.5  & 27.80 & 26.38 & 34.59 & 26.47 & 23.78 & 52.96 & 52.40 & 62.48 & 44.00 \\
GLM-5.2 & 29.52 & 26.69 & 34.97 & 31.70 & 24.74 & 56.37 & 58.21 & 42.08 & 68.81 \\
DeepSeek-V4.1-Flash & 36.36 & 34.31 & 35.56 & 40.69 & 34.86 & 62.92 & 60.72 & 64.86 & 63.19 \\
\midrule
\multicolumn{10}{l}{\textit{Official PostTrainBench v1.1 Results}} \\
GPT-5.5 & 27.23 & 25.28 & 29.60 & 27.48 & 26.55 & -- & -- & -- & -- \\
Claude Opus 4.7 & 28.56 & 30.08 & 31.01 & 26.69 & 26.43 & -- & -- & -- & -- \\
Kimi K3 & 31.96 & 30.03 & 41.03 & 30.30 & 26.49 & -- & -- & -- & -- \\
Claude Opus 4.8 & 33.84 & 29.16 & 40.73 & 34.73 & 30.76 & -- & -- & -- & -- \\
\midrule
\multicolumn{10}{l}{\textit{Reference Results}} \\
Base & 7.53 & 6.66 & 14.34 & 4.52 & 4.60 & 0.00 & 0.00 & 0.00 & 0.00 \\
Official Instruct & 51.14 & 49.41 & 63.75 & 44.81 & 46.58 & 100.00 & 100.00 & 100.00 & 100.00 \\
\bottomrule
\end{tabular}
}
\end{table*}

\paragraph{Fine-tuning substantially improves automated post-training.}
Fine-tuning on \tool{} trajectories consistently improves both model scales across the full and Lite settings. On PostTrainBench, RSI-Dojo-9B increases from 7.72 to 16.46, obtaining an 8.74-point gain, while RSI-Dojo-35B-A3B improves from 13.08 to 29.95, for a 16.87-point gain. Both fine-tuned models outperform their corresponding untrained models across all target base-model families. The improvements are even more significant on PostTrainBench-Lite: RSI-Dojo-9B rises from 0.56 to 15.71, while RSI-Dojo-35B-A3B increases from 5.80 to 53.29. Overall, these results demonstrate that training on the collected trajectories from \tool{} substantially enhances the models’ ability to construct effective post-training solutions across diverse base models and evaluation budgets.

\begin{wraptable}{r}{0.5\columnwidth}
\vspace{-0.8\baselineskip}
\centering
\caption{Model iteration statistics on PostTrainBench.}
\label{tab:iteration-statistics}
\setlength{\tabcolsep}{2pt}
\scriptsize
\resizebox{\linewidth}{!}{%
\begin{tabular}{lrrr}
\toprule
\multirow{2}{*}{Model} & At Least One & \multirow{2}{*}{Improved} & \multirow{2}{*}{Eff./Total (Rate)} \\
& Successful Run & & \\
\midrule
Qwen3.5-9B & 10/28 & 1/28 & 2/13 (15.38\%) \\
RSI-Dojo-9B & 24/28 & 15/28 & 29/247 (11.74\%) \\
Qwen3.6-35B-A3B & 15/28 & 11/28 & 17/68 (25.00\%) \\
RSI-Dojo-35B-A3B & 28/28 & 26/28 & 61/175 (34.86\%) \\
\midrule
GPT-5.4 & 28/28 & 23/28 & 112/586 (19.11\%) \\
GPT-5.5 & 28/28 & 27/28 & 108/522 (20.69\%) \\
GLM-5.2 & 28/28 & 26/28 & 57/140 (40.71\%) \\
DeepSeek-V4.1-Flash & 28/28 & 27/28 & 98/202 (48.51\%) \\
\bottomrule
\end{tabular}
}
\vspace{-0.6\baselineskip}
\end{wraptable}

\paragraph{RSI-Dojo-35B-A3B is competitive with frontier models on PostTrainBench.}
RSI-Dojo-35B-A3B achieves 29.95 on PostTrainBench, slightly exceeding GLM-5.2 at 29.52, GPT-5.5 at 27.80, and GPT-5.4 at 26.21. Although its distillation source models are not as strong as GLM-5.2, the final trained model achieves competitive performance with GLM-5.2 on this benchmark. On PostTrainBench-Lite, RSI-Dojo-35B-A3B reaches 53.29, comparable to GPT-5.5 at 52.96 and below GLM-5.2 and DeepSeek-V4.1-Flash at 56.37 and 62.92, respectively. Compared to the official leaderboard results, its performance also surpasses GPT-5.5 and Claude Opus 4.7. These comparisons show that the trajectories collected from \tool{} substantially narrow the gap between an open-weight 30B-level model and frontier 700B+ agents.

\paragraph{Fine-tuning improves execution reliability and iteration quality.}
Table~\ref{tab:iteration-statistics} shows the detailed training iteration statistics of each model. At Least One Successful Run counts settings in which the agent successfully completes at least one training, Improved counts final scores above the corresponding base score, and Eff./Total sums effective iterations which achieves local improvement and total iterations over completed runs. Although with the reprompted strategy and skills in AutoTrainess, Qwen3.5-9B and Qwen3.6-35B-A3B complete valid training runs for only 10 and 15 of the 28 evaluation settings, respectively. Their trajectories frequently fail while constructing executable training scripts or configuring the training pipeline; even completed runs seldom yield performance gains. Fine-tuning increases completion coverage to 24/28 for RSI-Dojo-9B and 28/28 for RSI-Dojo-35B-A3B, while increasing the number of settings with final improvement from 1 to 15 and from 11 to 26, respectively. RSI-Dojo-35B-A3B achieves 61 effective iterations out of 175, corresponding to an effective-iteration rate of 34.86\%. This is substantially higher than GPT-5.4 at 19.11\% and GPT-5.5 at 20.69\%, while requiring considerably fewer total iterations. Nevertheless, GLM-5.2 and DeepSeek-V4.1-Flash remain more iteration-efficient, with rates of 40.71\% and 48.51\%, respectively, suggesting further room to improve planning quality and experimental selection. 

\subsection{Ablation Study}
\label{sec:ablation-study}

\begin{figure*}[t]
    \centering
    \includegraphics[width=\textwidth]{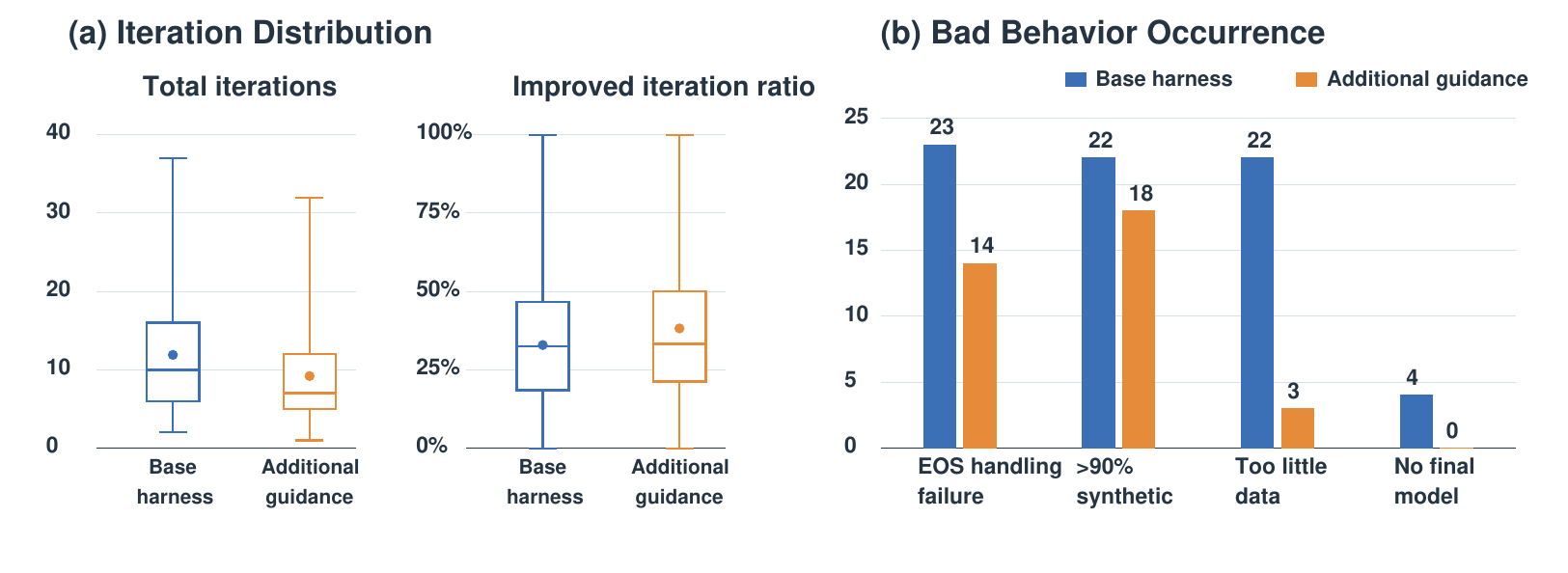}
    \caption{Effect of additional guidance on rollout data quality. Left: distributions of total iterations and the fraction of iterations that improve the best observed score. Right: counts of four common undesirable behaviors observed during trajectory collection.}
    \label{fig:additional-guidance-ablation}
\end{figure*}

\paragraph{Effect of additional guidance.}
We randomly sample 100 trajectories collected with the base harness and 100 trajectories collected after adding our targeted guidance, then analyze their quality. The additional guidance increases the trajectory retention rate from 42\% to 56\%. As shown in Figure~\ref{fig:additional-guidance-ablation}, it also makes the rollout process more efficient: the average number of iterations decreases from 11.93 to 9.25, while the average number of iterations that yield an actual improvement increases from 3.05 to 3.32. The additional guidance further reduces all four undesirable behaviors considered in our analysis. Cases where the trained model fails to terminate correctly because of improper EOS handling decrease from 23 to 14, training data dominated by synthetic data decrease from 22 to 18, trajectories using too little training data decrease from 22 to 3, and failures to provide a final model decrease from 4 to 0. These results show that guidance targeting recurrent failure patterns improves trajectory quality beyond the base harness.

\begin{table*}[t]
\centering
\caption{Ablation studies of trajectory filtering strategies on PostTrainBench-Lite.}
\label{tab:data-filtering-ablation}
\setlength{\tabcolsep}{5.5pt}
\renewcommand{\arraystretch}{1.12}
\small
\begin{tabular}{@{}lrrrrrrr@{}}
\toprule
Method & Overall & AIME 2025 & BFCL & GSM8K & HumanEval & Improved & \shortstack{At Least One\\Successful Run} \\
\midrule
Behavior only & 30.66 & 0.00 & 46.23 & 47.50 & 28.90 & 7/12 & 12/12 \\
Outcome only & 26.13 & \textbf{4.12} & 32.62 & 39.88 & 27.89 & 8/12 & 10/12 \\
\textbf{Both filters} & \textbf{53.29} & 0.00 & \textbf{95.05} & \textbf{69.49} & \textbf{48.62} & \textbf{9/12} & \textbf{12/12} \\
\bottomrule
\end{tabular}
\end{table*}

\paragraph{Effect of data filtering.}
We validates the two trajectory filtering strategies by training Qwen3.6-35B-A3B on two alternative subsets. In As shown in table~\ref{tab:data-filtering-ablation}, behavior-only filtering achieves an overall score of 30.66, while outcome-only filtering improves more settings, 8/12 versus 7/12, but obtains a lower score of 26.13. In contrast, applying both filters leads to a much stronger overall score of 53.29, showing a large gap compared with either single-filter variant. Both behavior-only filtering and the combined strategy complete at least one successful training run in all 12 settings, whereas outcome-only filtering succeeds in 10/12 settings. This indicates that the both filtering strategies are important for training agents that can reliably conduct post-training.

\subsection{Agent Behavior Analysis}
\label{sec:more-analysis}

\paragraph{Comparison across different models.}
By inspecting the training process of each model, we find that different models exhibit distinct post-training strategies. For example, GPT-5.4 favors many small step experiments, including LoRA on a few hundred examples and model merging. This produces long trajectories with few effective iterations, as shown in Table~\ref{tab:iteration-statistics}. It also rarely uses full-parameter training and often leaves the EOS configuration unchanged, leading to overly long responses in trained model. RSI-Dojo-35B-A3B, GLM-5.2, and DeepSeek-V4.1-Flash complete the 28 settings with substantially fewer iterations. The usage of distillation during training also varies in different models: GPT-5.4 and GPT-5.5 never use a larger teacher, compared with 2 runs for GLM-5.2 and 10 for DeepSeek-V4.1-Flash. Six of the DeepSeek-V4.1-Flash runs occur on HealthBench or Arena-Hard, with the others on GPQA Main, HumanEval, or AIME 2025. The official Claude Fable 5 traces also use distillation but only on HealthBench and Arena-Hard. These benchmarks are particularly amenable to teacher supervision because their evaluation targets are more complex. Although PosttrainBench does not prohibit using another local model for distillation~\cite{posttrainbench2026v11,posttrainbench2026v12policy}, this practice may allow a model to boost its benchmark score through a relatively simple strategy.

\begin{figure*}[t]
    \centering
    \begin{minipage}[t]{0.40\textwidth}
        \vspace{0pt}
        \centering
        \includegraphics[width=\linewidth]{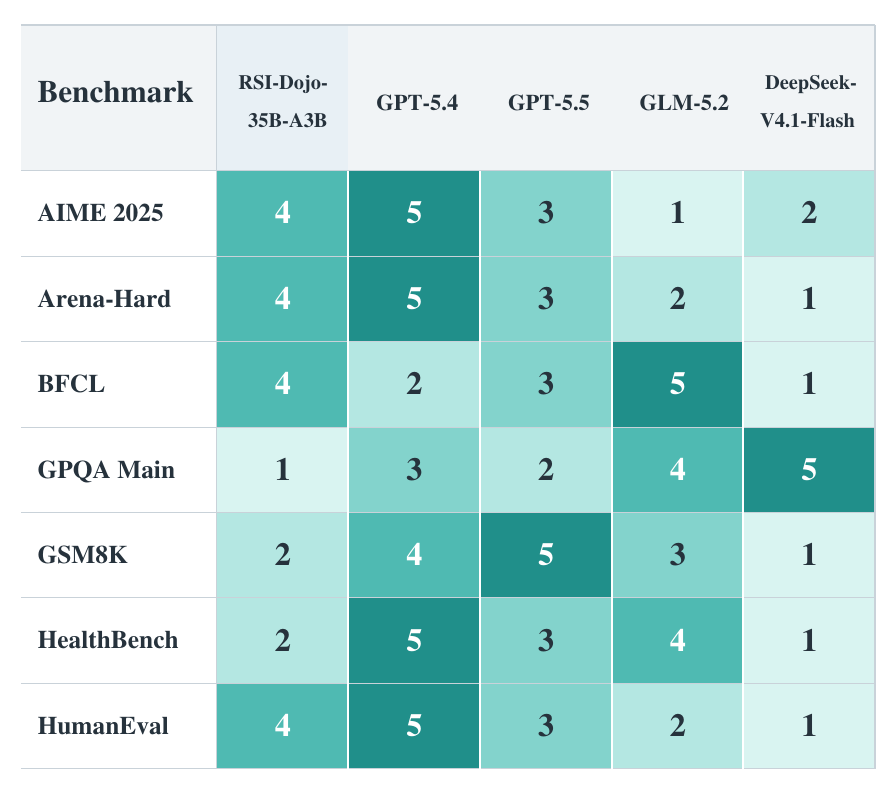}
        \captionof{figure}{Per-benchmark ranks among five models on PostTrainBench.}
        \label{fig:benchmark-rankings}
    \end{minipage}
    \hfill
    \begin{minipage}[t]{0.56\textwidth}
        \vspace{0pt}
        \centering
        \includegraphics[width=\linewidth]{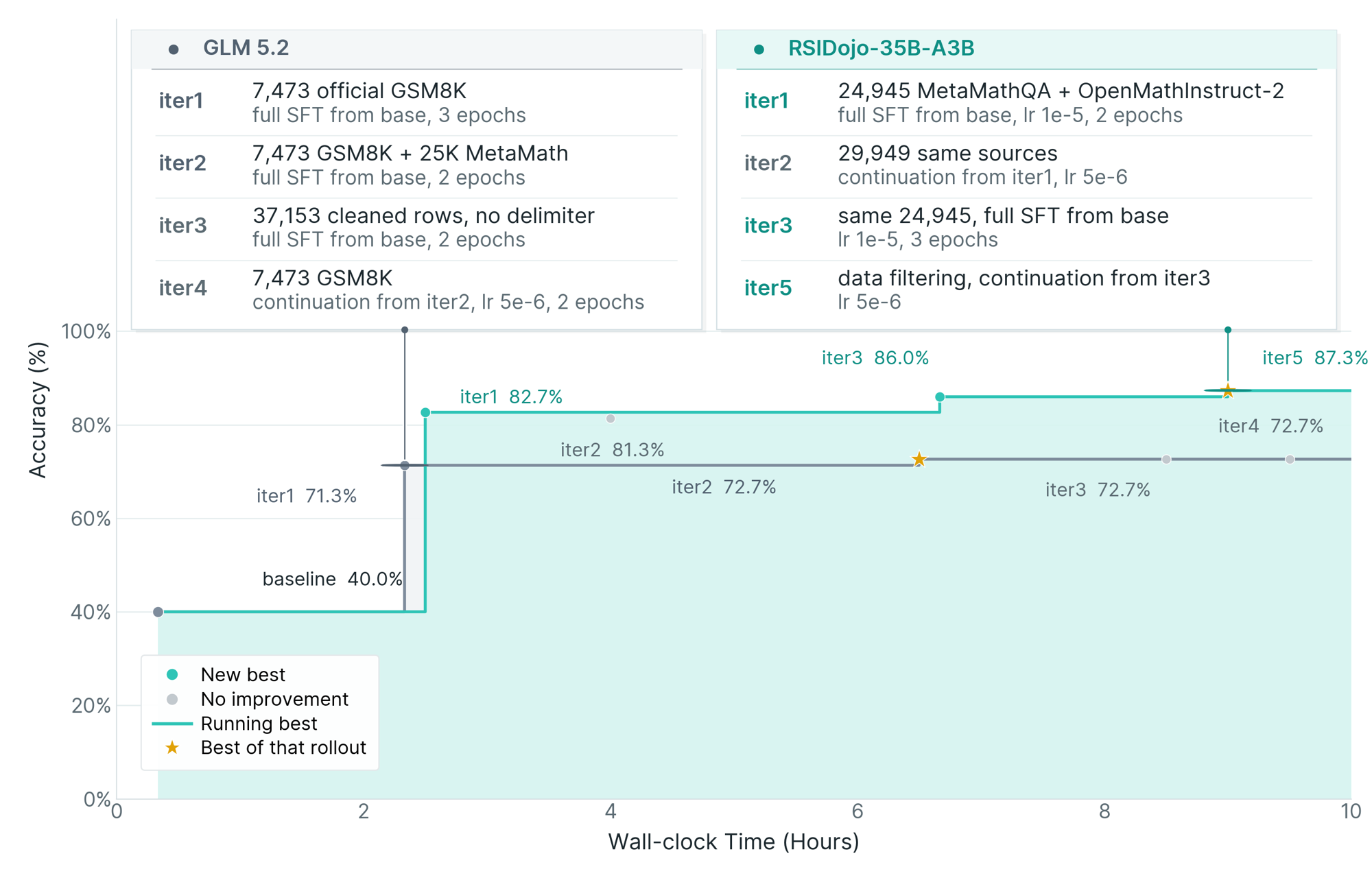}
        \captionof{figure}{An illustrative example shows the differences in training processes on GSM8K.}
        \label{fig:case}
    \end{minipage}
\end{figure*}

\paragraph{Comparison across different target benchmarks.}
Figure~\ref{fig:benchmark-rankings} shows that RSI-Dojo-35B-A3B ranks first on GPQA Main and second on GSM8K and HealthBench. It is competitive with GPT-5.4, GPT-5.5, and GLM-5.2 on BFCL and HumanEval, but weaker on AIME 2025 and Arena-Hard. The \tool{} training set includes relatively large proportions of math, general-QA, and medical tasks, which helps its strong performance on GSM8K, GPQA Main, and HealthBench. Although \tool{} also contains many difficult mathematical reasoning tasks, their rollouts rarely produce measurable improvements and are therefore filtered out, leaving limited supervision for AIME 2025. Arena-Hard exposes a different gap: substantial gains often rely on teacher distillation, but such trajectories are absent from our training data. This prevents RSI-Dojo from learning the strategy that produces large improvements on this benchmark.

\paragraph{Optimization patterns learned by RSI-Dojo.}
In trajectories with successful improvement, early gains typically come from correcting the training and output format. Full-parameter SFT and proper EOS token configuration prevent the termination failures often observed with LoRA or unchanged EOS settings. The dataset selection strategy is also crucial to the model’s results. In the GSM8K example in Figure~\ref{fig:case},  RSI-Dojo selects the more suitable \texttt{MetaMathQA} and \texttt{OpenMathInstruct-2} data, then improves further through data filtering and continued training at a lower learning rate. In later iterations, the agent often analyzes failed cases, refines the training subset, and continue training model with annealing. The evaluation target should evolve with the training stage. Evaluating on a small subset is useful in early stage. However, once subset performance saturates, continuing to optimize for it may favor local gains over benchmark-wide improvements. Therefore, high-performing trajectories often return to the complete evaluation set in later stages to align with the final objective.

\section{Conclusion}

In this paper, we introduce \builder{}, an automatic pipeline for converting research artifacts into executable model-training environments, and use it to construct \tool{} with 281 post-training tasks. Using \tool{}, we employ a trajectory collection method with an improved harness to collect high-quality trajectories, then apply quality filtering before training automated post-training behavior into Qwen3.5-9B and Qwen3.6-35B-A3B. The resulting models improve over their base models by 8.74 and 16.87 points on PostTrainBench, respectively, with RSI-Dojo-35B-A3B reaching 29.95 on PostTrainBench and 53.29 on PostTrainBench-Lite. These results demonstrate that training on high-quality post-training trajectories can strengthen a model's ability to autonomously construct data, optimize training recipes, diagnose issues, and iteratively improve the trained models.

\section{Limitations and Future Work}


Our current environments primarily cover PostTrainBench-style tasks. They do not yet represent the broader range of agentic post-training scenarios in which models must conduct longer-horizon research and adapt their data strategies to richer environment feedback. Future work will extend the task distribution to more realistic agentic settings such as RSIBench-Data \cite{meng2026rsibenchdata}.

Our current experiments also stop short of recursive self-improvement of the trained model itself. We internalize model-training capabilities with RSI-Dojo and evaluate whether it can improve smaller target models, but we do not close the loop by using our trained models to train an improved successor of themselves. After further strengthening its model training capabilities, we plan to explore and validate this stronger form of recursive self-improvement.

The current evaluation agent harness also limits the tools available during trajectory collection and evaluation. In Codex, \texttt{web\_search} is exposed as a custom tool and cannot be used by our self-hosted models, so we disable it throughout data collection to keep the tool configuration consistent. In the future work, we will integrate it back through an MCP server and enable the models to develop stronger data construction capabilities for harder post-training tasks.



\bibliography{iclr2026_conference}
\bibliographystyle{iclr2026_conference}

\clearpage

\end{document}